\documentclass[conference]{IEEEtran}
\usepackage{times}
\usepackage{soul}
\usepackage{url}
\usepackage{amsfonts}
\usepackage[hidelinks]{hyperref}
\usepackage[utf8]{inputenc}
\usepackage[small]{caption}
\usepackage{graphicx}
\usepackage{amsmath}
\usepackage{amsthm}
\usepackage{booktabs}
\usepackage{algorithm}
\usepackage{algorithmic}
\usepackage{amsmath}
\usepackage{amssymb}
\usepackage{subcaption}
\usepackage{tabularx}
\usepackage{multirow}

\newcommand{\pmem}{Optane-PM}

\newcommand{\code}[1]{\begin{ttcodefont}#1\end{ttcodefont}}

\title{Non-Volatile Memory Accelerated Geometric Multi-Scale Resolution Analysis\\[1.2ex]
    {\normalfont\large
        Andrew Wood\IEEEauthorrefmark{1}, Moshik Hershcovitch\IEEEauthorrefmark{2}, Daniel Waddington\IEEEauthorrefmark{2}, Sarel Cohen\IEEEauthorrefmark{3} \\Meredith Wolf\IEEEauthorrefmark{4}, Hongjun Suh\IEEEauthorrefmark{5}, Weiyu Zong\IEEEauthorrefmark{1}, Peter Chin\IEEEauthorrefmark{1}
    }\\[-1.5ex]
}

\author{
\IEEEauthorblockA{\IEEEauthorrefmark{1}Boston University \\ aewood@bu.edu\\spchin@bu.edu\\willzong@bu.edu}
\and
\IEEEauthorblockA{\IEEEauthorrefmark{2}IBM Research\\ moshikh@il.ibm.com\\daniel.waddington@ibm.com}
\and
\IEEEauthorblockA{\IEEEauthorrefmark{3}Hasso Plattner Institute\\sarel.cohen@hpi.de}
\and
\IEEEauthorblockA{\IEEEauthorrefmark{4}Williams College\\ mgw2@williams.edu}
\and
\IEEEauthorblockA{\IEEEauthorrefmark{5}Seoul National University\\hjsuh319@snu.ac.kr}
}

\begin{document}
\maketitle

\begin{abstract}
Dimensionality reduction algorithms are standard tools in a researcher's toolbox. Dimensionality reduction algorithms are frequently used to augment downstream tasks such as machine learning, data science, and also are exploratory methods for understanding complex phenomena. For instance, dimensionality reduction is commonly used in Biology as well as Neuroscience to understand data collected from biological subjects. However, dimensionality reduction techniques are limited by the von-Neumann architectures that they execute on. Specifically, data intensive algorithms such as dimensionality reduction techniques often require fast, high capacity, persistent memory which historically hardware has been unable to provide at the same time. In this paper, we present a re-implementation of an existing dimensionality reduction technique called Geometric Multi-Scale Resolution Analysis (GMRA) which has been accelerated via novel persistent memory technology called Memory Centric Active Storage (MCAS). Our implementation uses a specialized version of MCAS called PyMM that provides native support for Python datatypes including NumPy arrays and PyTorch tensors. We compare our PyMM implementation against a DRAM implementation, and show that when data fits in DRAM, PyMM offers competitive runtimes. When data does not fit in DRAM, our PyMM implementation is still able to process the data.
\end{abstract}

\section{Introduction}
Dimensionality reduction is a set of algorithms designed to extract the ``useful'' information that is often assumed to be embedded in a higher dimensional signal. Dimensionality reduction techniques have been used with great success in the field of machine learning~\cite{van2009dimensionality,sakurada2014anomaly} and information retrieval~\cite{tsuge2001dimensionality,venna2010information,kowalski2007information}. Some of the most popular variants, such as the linear Singular Value Decomposition (SVD)~\cite{golub1971singular} (and by extension Principal Component Analysis (PCA)~\cite{ringner2008principal}), as well as nonlinear versions such as ISOMAP~\cite{balasubramanian2002isomap} have even become standard items in a researcher's toolbox. Most recently, modern machine learning has pivoted to be more ``biologically'' based: developing algorithms and models which ever increasingly mimic the brain. Likewise, dimensionality reduction techniques have followed the same trend. One technique based on the behavior of the cortex called Geometric Multi-Scale Resolution Analysis (GMRA)~\cite{allard2012multi}, learns nonlinear manifolds in data by cleverly examining the local geometry of points in the pointcloud.

However, dimensionality reduction algorithms such as GMRA have historically been limited by the von-Neumann architectures they execute on: primarily constrained by small system memory (DRAM) capacity. In contrast, modern research has become dominated by ``big data'', with datasets such as ImageNet~\cite{deng2009imagenet}, Cifar100~\cite{krizhevsky2009learning}, MusicNet~\cite{thickstun2018invariances,trabelsi2017deep}, WMT14~\cite{axelrod2011domain} being exponentially larger than any in history.


In the last few years, a notable hardware breakthrough has been the invention of Intel Optane Persistent Memory Modules\footnote{Herein termed \pmem}. \pmem in particular communicates via the memory bus, circumventing bottlenecks such as PCIe lanes, using the same interface to the CPU as DRAM. While there are other Persistent Memory technologies, \pmem is the most mature product on the market. \pmem is based on 3D XPoint (3DXP) technology and operates at a cache-line granularity with a latency of around 300ns~\cite{spectra2020, izraelevitz2019basic}. While this latency is slower than current DRAM ({\raise.17ex\hbox{$\scriptstyle\mathtt{\sim}$}}100ns), it is 30x faster than the current state of the art NVMe SSDs. Additionally, a single DIMM of \pmem can reach 512GB, which is 8x larger than the available DRAM.  Thus, the maximum \pmem capacity of a commodity 2U server machine is 12TB - significantly more than DRAM.



By default, \pmem can be used in three modes: \textit{block device mode} in which it acts as a faster storage device, \textit{Memory Mode} in which it acts as high capacity volatile memory cached by DRAM, and \textit{App Direct} mode which allows load/store access to the device which is persistent. For a program to use \pmem in block device mode, the program must still pay an OS I/O penalty to access data on the device. In Memory Mode, a program does not pay this penalty, but must copy-off to additional storage devices if persistence is required. App Direct mode treats the memory as separate from volatile memory in the system.  \pmem is exposed as a DAX (direct access) device.  Once it has been mapped into the process virtual address space, it can be accessed directly through load/store.  Under-the-hood, data is moved around in cache lines and 256B blocks internally to the \pmem device.  Although App Direct mode is very powerful, it does require programs to be aware of this "special" memory and pay close attention to crash-consistency and data recovery (normally requiring access to low level persistent memory and software transaction support libraries).

While \pmem technology is still in its infancy, there are some early adopters in databases (see \cite{SAPHANA}).  However, to the best of our knowledge, we are the first to exploit this technology to accelerate dimensionality reduction algorithms such as GMRA. 

In this paper, we use components of MCAS~\cite{mcas2021}, a key-value store framework built from the ground up for persistent memory, to implement GMRA. Specifically, we use a simplified MCAS library integrated into the Python programming language called Python Micro-MCAS (PyMM)~\cite{WaHeDi21PyMM}. PyMM uses local persistent memory resources only, unlike the broader MCAS solution which is a network-attached programmable key-value store.

One of the key features of PyMM is its support for native Python types such as NumPy \code{ndarray} and PyTorch \code{tensor}. We use PyMM to re-implement GMRA. In total, we were required to only 15 lines of code change (out of $1000+$ lines of code) to integrate our existing implementation with PyMM. We then show that despite the higher latency of \pmem, PyMM-GMRA is still comparable to its DRAM only counterpart when data fits into DRAM. Additionally, short of writing your own Python interface to persistent memory, PyMM provides a straightforward solution when data is larger than available DRAM. Specifically, we evaluate our performance on the MNIST~\cite{lecun1998mnist} and Cifar10~\cite{krizhevsky2009learning} datasets.

\section{Background and Related Work}

\subsection{MCAS: Memory Centric Active Storage}

\begin{figure}[t]
    \centering
    \includegraphics[width=\linewidth]{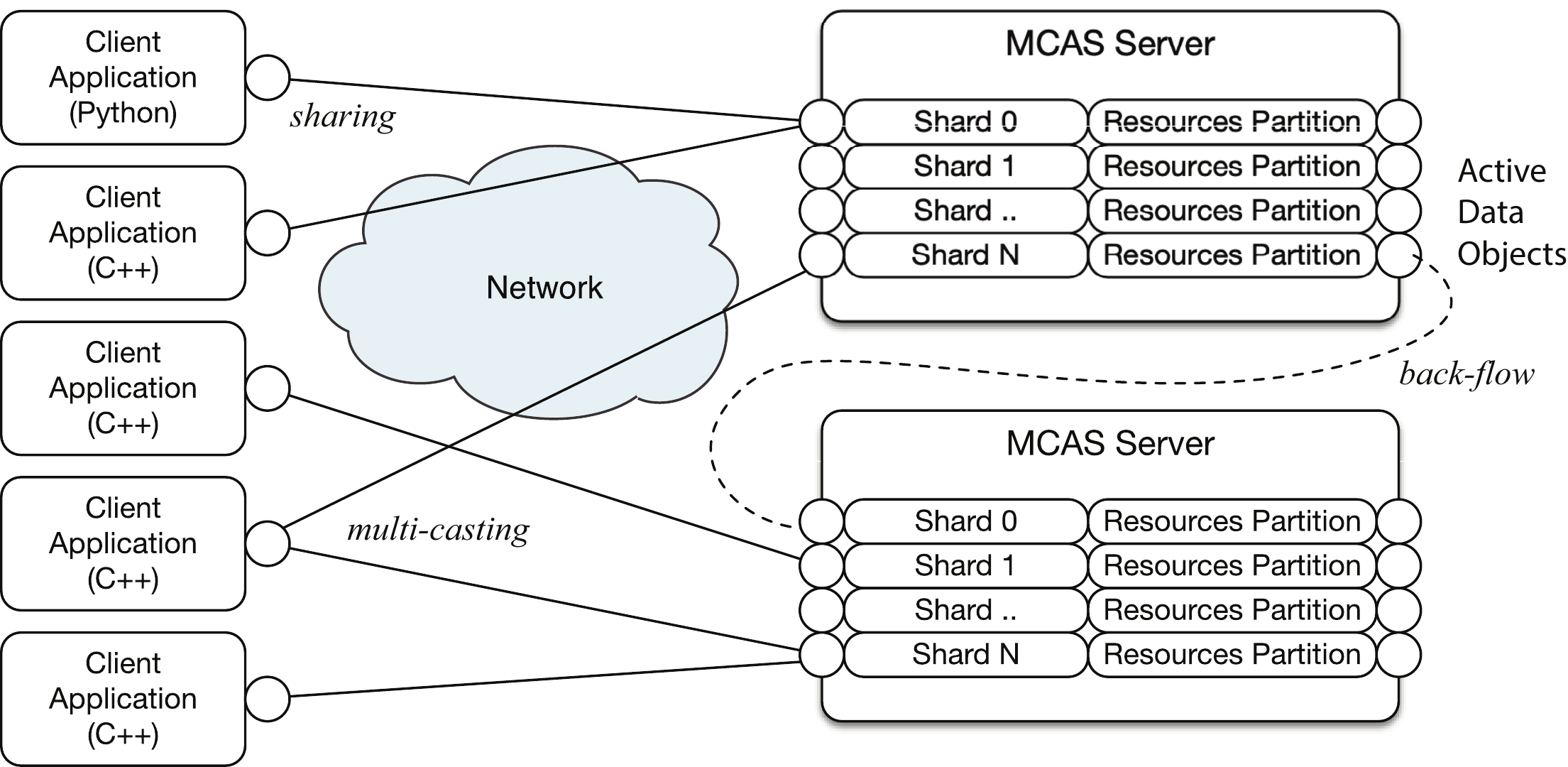}
    \caption{The high level architecture of MCAS. MCAS abstracts Optane in DAX mode to a key-value store api via the network which allows clients to load/store memory via the system memory bus while maintaining data persistence. Image taken from~\cite{mcas2021}.}
    \label{fig:mcas_arch}
\end{figure}

MCAS is a networked key-value store designed to interface with Optane devices in DAX mode. While the MCAS architecture can be seen in Figure~\ref{fig:mcas_arch}, it can generally be thought of as an abstraction layer between the persistent memory and a client application. A key feature of MCAS is the support of in-store operations. This feature, called Active Data Objects (ADO)s, allows code to run directly on the MCAS server without needing to transfer values to the client or move data objects into DRAM. Thus, MCAS enabled systems allow client applications to modify existing data objects, create new data objects, and delete data objects without the need to pay runtime penalties for copying/transferring data. Additionally, if persistence is enabled, all data objects will live after program execution and after system reboots. Therefore, programs can avoid paying additional I/O runtime penalties by using MCAS. We note that programs will also need to add crash-consistency code to ensure data recovery in the event of a system crash, this behavior is not enabled by default.

MCAS currently supports pure Python ADOs which execute Python code directly on the MCAS server. However, this api is not easy to use, and requires considerable skill by the developers to write correct applications. Therefore, a novel abstraction library called PyMM was developed to encapsulate MCAS Python calls as well as support native Python types.





\begin{figure}
    \centering
    \includegraphics[width=\linewidth]{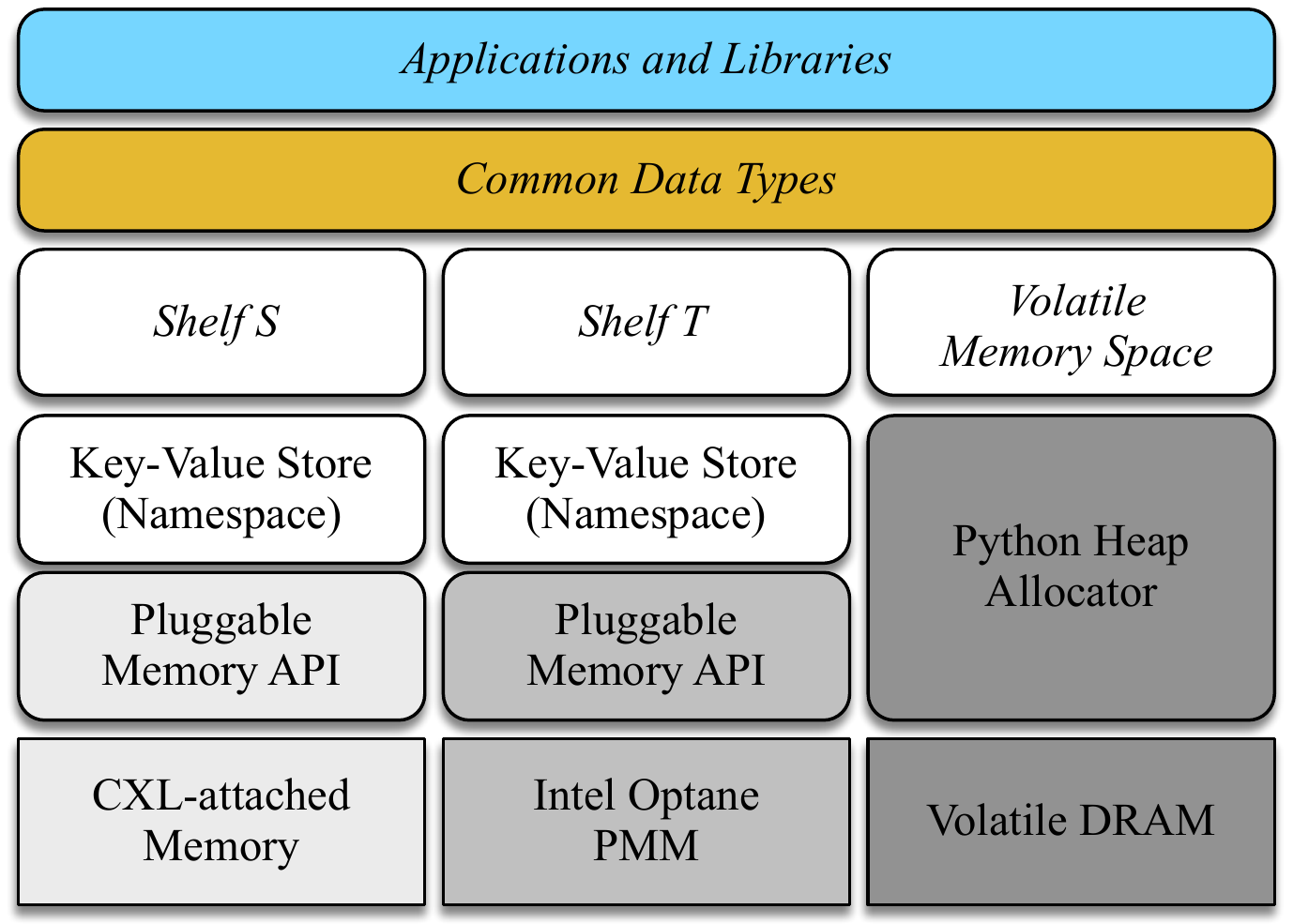}
    \caption{The PyMM architecture. PyMM exports micro-MCAS, a networkless version of MCAS, to the Python programming language, and also provides support for native Python types as well as NumPy arrays and PyTorch tensors. PyMM abstracts away memory management from the client through the use of a shelf datatype which can be seen above.}
    \label{fig:pymm_stack}
\end{figure}

\subsection{PyMM: Python Micro-MCAS}
As previously mentioned, PyMM is a new Python library that encapsulates the MCAS api and provides a significantly easier interface. Specifically, PyMM is built on top of a networkless MCAS implementation which is called Micro-MCAS. Additionally, PyMM includes built-in support for native Python types such as NumPy arrays and PyTorch tensors. While arbitrary datatypes can be stored on an MCAS server, the client is responsible for writing the code that interfaces their datatypes with the persistent memory such as writing metadata, etc. For NumPy and PyTorch, most linear algebra operations would require custom code. PyMM handles this for the client, allowing the client to use existing NumPy algorithms as-is, and without major modification to their existing NumPy/PyTorch code. The PyMM stack can be seen in Figure~\ref{fig:pymm_stack}.

PyMM is built around a \textit{shelf} datatype. A shelf represents a logical grouping of data objects which are held in a specific set of memory resources. In our case, the memory resource is Optane acting in DAX mode, enabling both high memory speeds and persistence. Therefore, PyMM shelf objects will exist across program executions and system resets. Data objects can also be exchanged between DRAM or any other memory device which is directly attached to the CPU via the memory bus, including GPUs. Shelf objects can be created in two modes: \emph{devdax} (visible as a character device file) or \emph{fsdax} (visible as block device files). In our case, we create shelf objects in fsdax mode.

In our experiments, we use PyMM's built-in support for NumPy arrays which correspond to the shelf type \texttt{pymm.ndarray}. This shelf type is polymorphic with NumPy's native \texttt{ndarray} type, meaning that all existing NumPy algorithms will work (and operate in-place) with a shelf.

For data-intensive algorithms such as dimensionality reduction, fast system memory with high capacity and persistence is required. For dimensionality reduction algorithms, the entire dataset is usually required to exist in memory at once, as global knowledge is needed. These algorithms commonly construct complicated internal representations of the data which they then use to perform dimensionality reduction. For instance, SVD and PCA require computing the gram matrix ($A^TA$ or $AA^T$), ISOMAP and LDA build diffusion maps, and GMRA uses approximate nearest neighbor strategies. These complicated data structures are expensive to build, and consume memory proportional to the number of data points. Therefore, dimensionality reduction techniques require persistence: they only build these structures once for a specific dataset, and then save that structure for future use/reuse. By using MCAS, the I/O penalty for serializing these complicated structure to disk can be avoided.

While there has been work on building these structures in a streaming fashion~\cite{rachakonda2016memory,huang2020memory,mitliagkas2013memory,ghashami2015frequent}, these algorithms do not provide general case solutions for all dimensionality reduction techniques, and they often are approximate algorithms~\cite{huang2020memory,ghashami2015frequent}. By using PyMM (and therefore MCAS), dimensionality reduction techniques are able to process large datasets that were not otherwise possible without using streaming approximations. By specifically using PyMM, these algorithms require the fewest changes to their existing codebases in order to benefit from MCAS.

We note that a separate project, called PMDK~\cite{PMDK} also provides persistent memory access to Optane via the memory bus. A C/C++ wrapper called PyNVM~\cite{PyNVM} provides PMDK functionality to Python and allows for custom datatypes to be compatible with PMDK. However, PyNVM does not use a subclassing strategy and also does not support flat datatypes such as NumPy ndarrays or PyTorch tensors. Thus, we chose to use PyMM and MCAS for our implementation.




\begin{figure}[!t]
\centering
\includegraphics[width=0.5\textwidth]{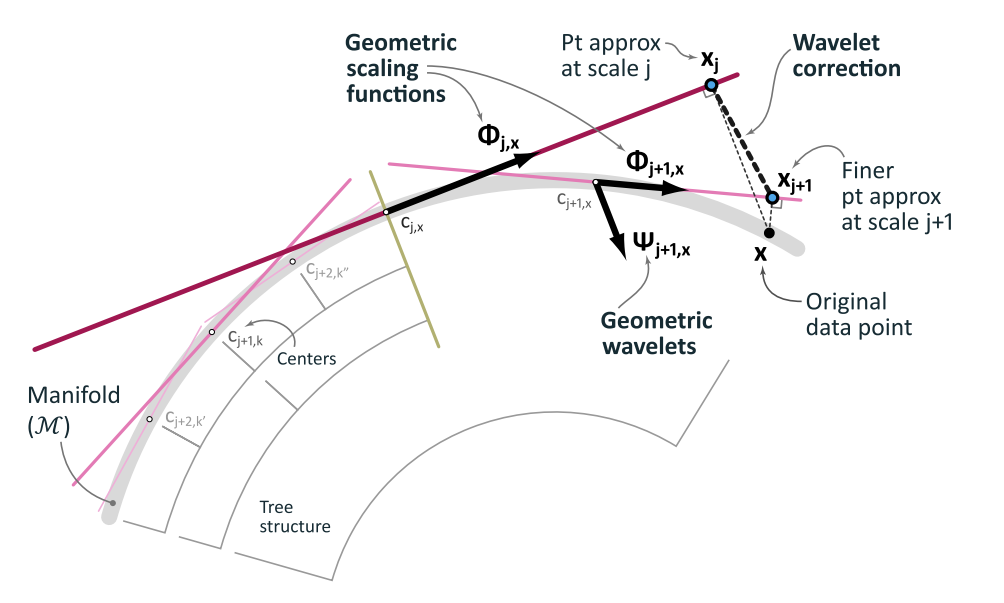}
\caption{A visualization of the linear approximation of a point using a basis function at scale $j$, the approximation at a finer scale $j$+1 and difference operator (geometric wavelet) from scale $j$ to scale $j+1$. This image was taken from \protect\cite{allard2012multi}.}
\label{fig:gmra}
\end{figure}

\subsection{Geometric Multi-Scale Resolution Analysis (GMRA)}
GMRA is a dimensionality reduction technique that is inspired by the cortex. At the microscale, neurons in the cortex fire, which induces synchrony between neurons. Neural synchrony at the macroscale produces patterns, which combined with other firings, have long been believed to be the intermediary representation of data~\cite{dicarlo2012does}. Macroscale firings can also be viewed as dimensionality reduction: at the macroscale, due to neural synchrony, at any given point only a subset of the neurons are firing. Therefore, by viewing ever larger populations of neurons, it is believed that the cortex computes lower-dimensional representation of data at different scales in increasing layers of abstraction. GMRA mimics this behavior by processing a point cloud at different scales to produce increasingly fine-grained manifolds.

The GMRA algorithm contains three steps to compute these manifolds at different scales:
\begin{enumerate}
    \item It computes a leveled tree decomposition of the manifold into \textit{dyadic cells}. Dyadic cells have the following properties:
        \begin{enumerate}
            \item Each dyadic cell contains a subset of the pointcloud, and the entire subset exists within a sphere with a fixed radius.
            \item The children of a dyadic cell divide the points contained in the parent into disjoint subsets.
            \item The children of a dyadic cell cover the points inside the parent.
        \end{enumerate}
        
        It is worth noting that due to these rules, each level of the tree covers the entire point cloud, and all cells at a level are disjoint. The root node of the tree contains the entire point cloud, and the final level of the tree contains the finest-scale (smallest radii) grouping of the points.

    \item It computes a $d$-dimensional affine approximation for each dyadic cell. This approximation represents the basis of each dyadic cell and is a linear piecewise approximation (i.e. the SVD decomposition of the cell's covariance).
    \item It computes a sequence of low-dimensonal affine difference operators that encode the difference between subsequent levels of the tree (i.e. scales). These difference operators allow efficient querying of the points by scale.
\end{enumerate}

GMRA essentially searches for manifolds by grouping points according to their local topology and produces piecewise linear approximations of the groups. First, GMRA considers the entire point cloud as a group and computes a linear approximation of the entire dataset using SVD~\cite{golub1971singular}. Then, it iteratively partitions the point cloud into smaller groups by finding open balls which contain the maximum number of points, where it computes linear approximations for those groups, producing the next level of cells in the tree. This cycle repeats until some halting criteria is met. To query the tree, GMRA computes difference operators between levels of the tree which allow a query to start at the root, lookup the embeddings for the query, and then walk the tree, updating the embeddings using the linear difference operators (a process which can be seen in Figure~\ref{fig:gmra}).

By computing the linear approximation of a cell using SVD of the covariance, the approximation geometrically becomes a parallel plane that passes through the mean of the cell and is approximately tangent to the local area of the manifold in the cell. As seen in Figure~\ref{fig:gmra}, the linear approximation, called the \textit{scaling function} fit to each group can be queried by starting at scale 0 (the roughest scale), getting the approximation for the query at that scale, and then applying the difference operator (\textit{wavelet correction}) to get the approximation at the next scale (finer scale). Since each level of the tree represents the decomposition of the point cloud at a scale, by walking from the root to the child at the appropriate level, GMRA produces low-dimensional embeddings for each point at arbitrary scales. GMRA has been used to improve performance of downstream tasks~\cite{tran2014geometric,chen2012fast} such as classification and anomaly detection.

\section{Our Approach}
\subsection{GMRA Implementation}
The GMRA tree decomposition step can be implemented in a few ways~\cite{allard2012multi}:
\begin{enumerate}
    \item Use approximate nearest neighbors to construct a weighted graph where the nodes represent points in the point cloud, and draw weighted edges from a node to its $k$ nearest neighbors. The weight of this edge takes the form $e^{\frac{||x_i - x_j||_2^2}{\sigma}}$ where $k$ and $\sigma$ are hyperparameters. In practice, $k$ is taken between 10 and 50, and $\sigma$ is typically the distance between $x_i$ and its $\left \lfloor{\frac{k}{2}}\right \rfloor$ nearest neighbor.
    
    By partitioning this weighted graph (e.g. METIS routines~\cite{karypis1998fast}), the dyadic cells are constructed as the vertex groups left after each separation, while the tree structure is given by the action of separation. For instance, the partitioning algorithm is run recursively on all previous partitions, yielding the dyadic cells and their parent-child relationships.

    \item Use the CoverTree algorithm~\cite{beygelzimer2006cover} to partition the data into a level tree based on radii at different scales. The CoverTree algorithm builds a graph where each node corresponds to a single point in the pointcloud, and draws a directed edge $x_i\rightarrow x_j$ if $x_j$ is within radius $2^s$ from $x_i$ (as measured by euclidean distance).
    
    Since a CoverTree has a root node, the dyadic cells are constructed by computing all vertices reachable within the radius $2^j$ at scale $j$ given the root of the subtree. The procedure then recurses for all children of the root reachable by radius $2^{j-1}$ for the next scale $j-1$. Note that in order to use the CoverTree construction, the max scale must be known apriori (to the construction) and can be an expensive operation. However, CoverTree construction is the only approach which provides the theoretical bounds for the GMRA algorithm.
    
    \item Iterated PCA. At the largest scale, compute the top $d$ principle components, and then assign the points into two children based on the sign of their $d+1$-st singular vector. The tree relationship is constructed by recursing on the two children.
    
    \item Iterated $k$-means clustering. At the largest scale, partition the data into $d$ clusters, and recurse on each cluster.
\end{enumerate}

Our implementation currently supports the CoverTree construction. The original implementation of GMRA is in MATLAB and uses custom C/C++ code which are distributed in binary form. This has the unfortunate side effect of locking the GMRA-MATLAB implementation to a specific, outdated version of MATLAB using outdated (and in some cases, unsupported) third party libraries. Additionally, GMRA-MATLAB will not interface with MCAS and therefore is constrained by DRAM.

Our implementation also makes use of custom C/C++ code, however our C/C++ is distributed in source code form and is exported to Python using libtorch (the C++ library of PyTorch~\cite{paszke2019pytorch}). Most importantly, our implementation interfaces with PyMM and therefore can leverage its PyTorch and NumPy support. Specifically, we implemented the tree construction using the CoverTree algorithm in C++ for speed and parallelism, and then exported our code to Python using Pybind11~\cite{jakob2017pybind11}. Final wavelet construction occurs in pure Python with full PyMM support. Our implementation demonstrated PyMM's seamless support for existing NumPy routines. For instance, we used NumPy's svd algorithm and scipy's qr algorithm with \code{pymm.ndarray} datatypes on persistent memory.

\begin{table}[!b]
\centering
\begin{tabular}{|c|c|c|c|}
\toprule
Name            & \# examples   & dimensionality    & \# labels \\
\midrule
MNIST       & 70k           & 784               & 10        \\
Cifar10     & 60k           & 3072              & 10        \\
\bottomrule
\end{tabular}
\caption{A breakdown of the datasets. Note that while Cifar10 has fewer examples, the dimensionality of each example is significantly higher.}
\label{table:datasets}
\end{table}

\begin{figure*}[!th]
    \begin{subfigure}{0.5\textwidth}
        \centering
        \includegraphics[width=0.6\linewidth]{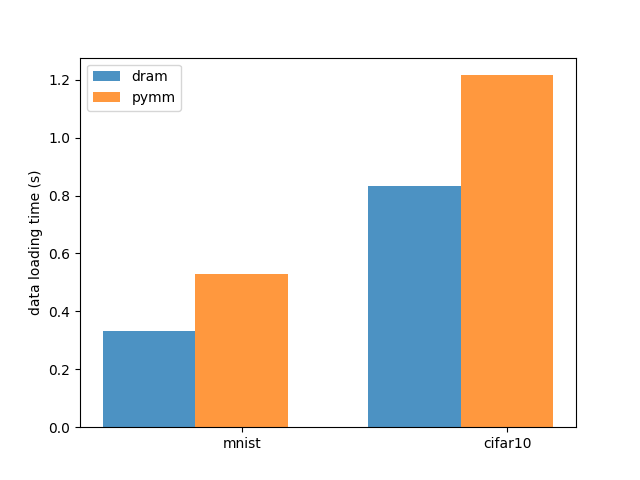}
        \label{fig:sfig1}
    \end{subfigure}%
    \begin{subfigure}{0.5\textwidth}
        \centering
        \includegraphics[width=0.6\linewidth]{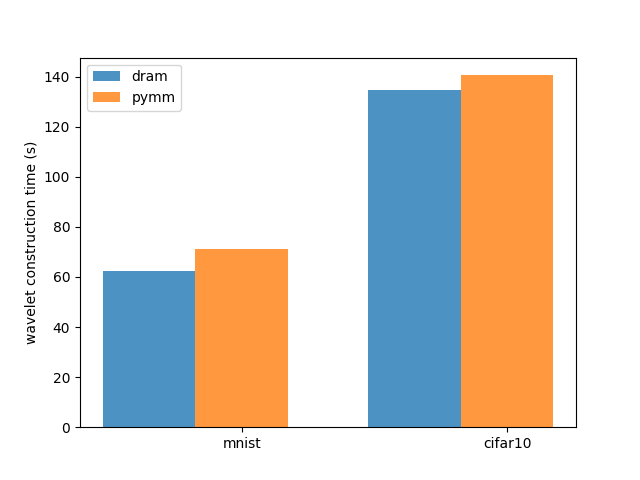}
        \label{fig:sfig2}
    \end{subfigure}
    \begin{subfigure}{0.5\textwidth}
         \centering
        \includegraphics[width=0.6\linewidth]{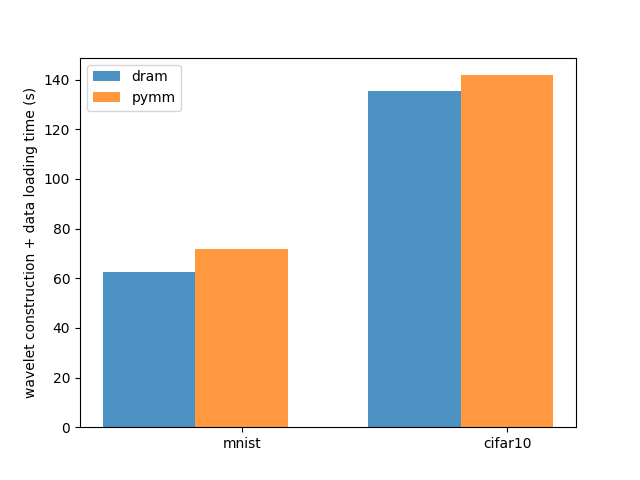}
        \label{fig:sfig3}
    \end{subfigure}%
    \begin{subfigure}{0.5\textwidth}
        \centering
        \includegraphics[width=0.6\linewidth]{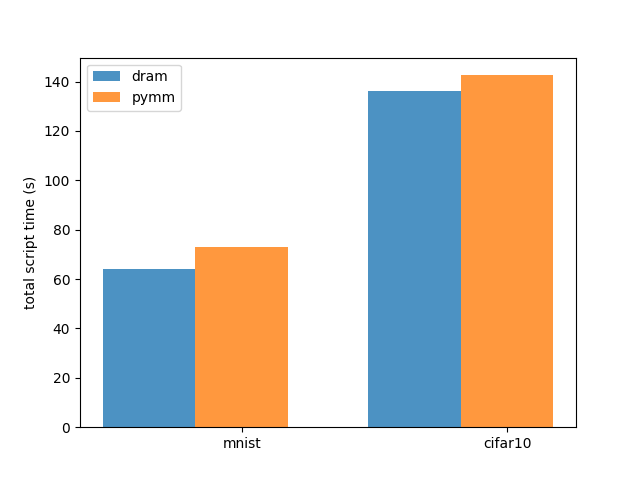}
        \label{fig:sfig4}
    \end{subfigure}
\end{figure*}

\begin{table*}[!t]
    \centering
    \begin{tabular}{|c|c|c|c|c|c|}
    \toprule
    Dataset & Data Loading DRAM (s) & Data Loading PyMM(s) & Wavelet Construction DRAM(s) & Wavelet Construction PyMM(s) & num trials \\
    \midrule
    MNIST & $0.33\pm 2.02\times10^{-3}$ & $0.53\pm 1.73\times10^{-3}$ & $62.4\pm 1.88$ & $71.2\pm 1.55$ & 20 \\
    Cifar10 & $0.83\pm 2.58\times10^{-3}$ & $1.2\pm 8.26\times10^{-3}$ & $134.7\pm 5.07$ & $140.6\pm 5.24$ & 20 \\
    \bottomrule
    \end{tabular}
    \caption{Top: Timing results of our experiments. The top two figures show the timings from loading data and wavelet construction respectively. The bottom two figures aggregate these results into the total time of using DRAM/PyMM and the total running time of the experiments respectively. Note that while PyMM performs much worse during the data loading phase, this phase comprises of ~1\% of the total time spent using PyMM/DRAM. Bottom: Expected timing values including standard deviations. MNIST took 209.35MB on the shelf and Cifar10 took 703.12MB on the shelf.}
    \label{tab:timing_results}
\end{table*}
\section{Experiments}
We took two datasets, MNIST~\cite{lecun1998mnist} and Cifar10~\cite{krizhevsky2009learning} for our experiments. Both datasets contain images, but with different number of examples, dimensionality, and content. Most notably the Cifar10 dataset contains larger, color images while the MNIST dataset contains smaller, grayscale images. A breakdown of the datasets (preprocessed and in original form) is shown Table~\ref{table:datasets}. 

For our experiments, we measured the runtime that the GMRA algorithm took to process both the MNIST and Cifar10 datasets. We computed the CoverTree structure outside of our expeiments, as it only needs to occur once, is shared between the PyMM and DRAM versions, and only uses DRAM. We then serialized the CoverTree to disk. To compare DRAM to PyMM, we ran the tree deserialization, Dyadic cell computation, and wavelet computation using both a DRAM and PyMM enabled version of GMRA. Since the PyMM api is similar to the raw NumPy api, and supports NumPy api operations on PyMM-stored arrays, the differences between the two implementations was minimal. Our experiments were run with PyMM crash consistency disabled.

We ran the partial GMRA algorithm (CoverTree deserialization through wavelet construction) 10 times each using DRAM and PyMM, and recorded timing splits for each stage of the script. We note that changing between DRAM and PyMM has minimal effect on the Dyadic cell construction (due to this operation always occurring in DRAM and being minorly influenced by L2-L4 cache misses). However, all major linear algebra operations occur during the wavelet construction stage, and this operation will either occur in DRAM or PyMM as well as storing the dataset itself. In our experiment, we included the cost of moving the dataset from DRAM/disk to PyMM in the PyMM reported times.

In our experiments, we used a server equipped with two Intel Xeon Gold 6248 processors containing 80cpu cores running at 2.5GHz base clock. The server is also equipped with 384GB of DDR4 DRAM and 786GB of Optane DC. 
This server also has a NVIDIA Tesla m60 GPU, however we did not make use of it in our experiments.




\section{Results and Discussion}
In our experiments, we observed that our PyMM GMRA implementation performed slower than the DRAM implementation. This behavior is expected, and results can be seen in Table~\ref{tab:timing_results}. We were however surprised at how fast the PyMM implementation was compared to how slow Optane memory is with respect to DRAM speeds. In our experiments, we observed a slowdown of 1.6x on MNIST and a 1.45x slowdown on Cifar10 for loading data into PyMM versus DRAM. This finding is surprising given the reported latency: Optane is 3x slower than DRAM. For wavelet construction; the stage where PyMM has the greatest influence over GMRA, we observed a slowdown of 1.14x on MNIST and a 1.04x slowdown on Cifar10. Our timing results can also be seen in Table~\ref{tab:timing_results}.

While the slowdown is almost negligible for MNIST and Cifar10, it is worth noting that for larger datasets, the slowdown can amount to hours of additional runtime. However, there is a hidden cost that we did not measure in our experiments: the penalty for serializing the GMRA wavelets to disk. This I/O penalty only applies to the DRAM version since PyMM is persistent, and it is unlikely that this serialization would close the runtime gap on large datasets. However, for downstream tasks, the I/O penalty of using a DRAM system would be nontrivial, since there is no guarantee that the lower dimensional dataset would fit into DRAM, and that read speeds from persistent storage such as a NVMe SSD or HDD would be significantly slower than load speeds from PyMM.

Our experiments had crash consistency disabled. Crash consistency is implemented via a software logging strategy in PyMM that occurs at every write operation. While crash consistency is important for our algorithms, it induces a copy operation every time a data object is written. Since this policy is in software, this adds significant overhead to the runtime of any algorithm which uses PyMM if crash consistency is enabled. We omitted crash consistency at this point because our algorithm is not crash consistent. In order to make our algorithm crash consistent, it would require additional code changes which would create an unfair comparison between our PyMM implementation and a (already crash inconsistent) DRAM implementation.


\section{Future Work}
One major improvement that we leave to future work is to integrate our custom C/C++ tree structures into PyMM. Currently, our tree structures (CoverTree and Dyadic cell tree) are implemented on DRAM and execute in series. In the future, we would like to put both of these structures on persistent memory using PyMM and operate on them in-place. A further direction is to add parallelization to our CoverTree implementation. CoverTrees are highly parallelizable, however we omitted this in our current implementation in order to simplify the code.

Additionally, we envision a scenario where GMRA and other dimensionality reduction techniques are supported by all the features available in MCAS. Currently, PyMM can only access local storage devices, which caps the capacity to how many \pmem drives can fit inside a single machine. Fully-fledged MCAS on the other hand provides a network abstraction, allowing memory to be shared across multiple machines. This would enable our algorithms to scale-out across nodes and process datasets currently only possible by large Hadoop clusters.

Our last avenue of future work is to utilize GPU resources. By transferring data between persistent memory and GPU memory, we can further accelerate our algorithms either in down stream tasks like using the lower-dimensional datasets for learning, or by using GPU resources in our algorithms to speed up calculations.  CXL-attached GPU and FPGA accelerators combined with persistent memory might also be an interesting direction for future research.

\section{Conclusion}
In conclusion, we have demonstrated that MCAS and PyMM provide a suitable framework to implement data intensive algorithms such as GMRA. While persistent memory has higher latency than DRAM when the dataset fits into DRAM, PyMM-GMRA still provides competitive runtimes. When data does not fit into DRAM, PyMM-GMRA is the only implementation which can process this data. We demonstrate that with PyMM, data intensive algorithms have new opportunities to escape the classical restrictions of von-Neumann architectures and can be used more effectively in the world of big data.

\bibliographystyle{IEEEtran}
\bibliography{IEEEabrv, main}

\end{document}